\pdfoutput=1
\documentclass[11pt]{article}

\usepackage{acl}

\usepackage{booktabs}
\usepackage{times}
\usepackage{latexsym}
\usepackage{graphicx}
\usepackage{multirow}
\usepackage{amsmath}
\usepackage{algorithm}
\usepackage{algorithmic}
\usepackage{tabularx}
\usepackage{amssymb}
\usepackage{amsfonts}
\usepackage[T1]{fontenc}
\usepackage[utf8]{inputenc}

\usepackage{microtype}

\usepackage{inconsolata}

\title{Plausible Extractive Rationalization \\through Semi-Supervised Entailment Signal}

\author{Yeo Wei Jie\textsuperscript{1}, Ranjan Satapathy\textsuperscript{2}, Erik Cambria\textsuperscript{1}\\
\textsuperscript{1}Nanyang Technological University \\
\textsuperscript{2}A*STAR, IHPC \\
\texttt{yeow0082@e.ntu.edu.sg} \\
}

\begin{document}
\maketitle
\begin{abstract}
The increasing use of complex and opaque black box models requires the adoption of interpretable measures, one such option is extractive rationalizing models, which serve as a more interpretable alternative. These models, also known as Explain-Then-Predict models, employ an explainer model to extract rationales and subsequently condition the predictor with the extracted information. Their primary objective is to provide precise and faithful explanations, represented by the extracted rationales. In this paper, we take a semi-supervised approach to optimize for the plausibility of extracted rationales. We adopt a pre-trained natural language inference (NLI) model and further fine-tune it on a small set of supervised rationales ($10\%$). The NLI predictor is leveraged as a source of supervisory signals to the explainer via entailment alignment. We show that, by enforcing the alignment agreement between the explanation and answer in a question-answering task, the performance can be improved without access to ground truth labels. We evaluate our approach on the ERASER dataset and show that our approach achieves comparable results with supervised extractive models and outperforms unsupervised approaches by $> 100\%$. Code available at \url{https://github.com/wj210/NLI_ETP}.
\end{abstract}

\section{Introduction}
Large language models such as Google's BERT~\cite{devlin2018bert} and OpenAI's GPT series~\cite{brown2020language} are gaining widespread adoption in natural language processing (NLP) tasks. These models achieved impressive performance in multiple NLP tasks ranging from solving text generation to information extraction~\cite{liu2023summary}. However, little is known regarding how answers are generated or which portion of the input text the model focuses on. These flaws highlight concerns surrounding trust and fear of undesirable biases in the model's reasoning chain. Explainable AI (XAI) is currently an active field of research aimed at addressing these issues~\cite{adadi2018peeking,camsur,yeo2023comprehensive}. In this work, we focus on extractive rationalizing models~\cite{lei2016rationalizing}, which are also known as Explain-Then-Predict (ETP) models, and are designed towards producing highlights serving as \textbf{faithful} explanations. Faithfulness is defined as serving an explanation that represents the model's reasoning process for a given decision, while plausibility refers to the level of agreement with humans~\cite{jacovi2020towards}. An advantageous characteristic of ETP models is that they concurrently produce the explanation and the task label, eliminating the necessity for an added layer of interpretation. 

\begin{figure}[h]
\centering
\includegraphics[width=\columnwidth]{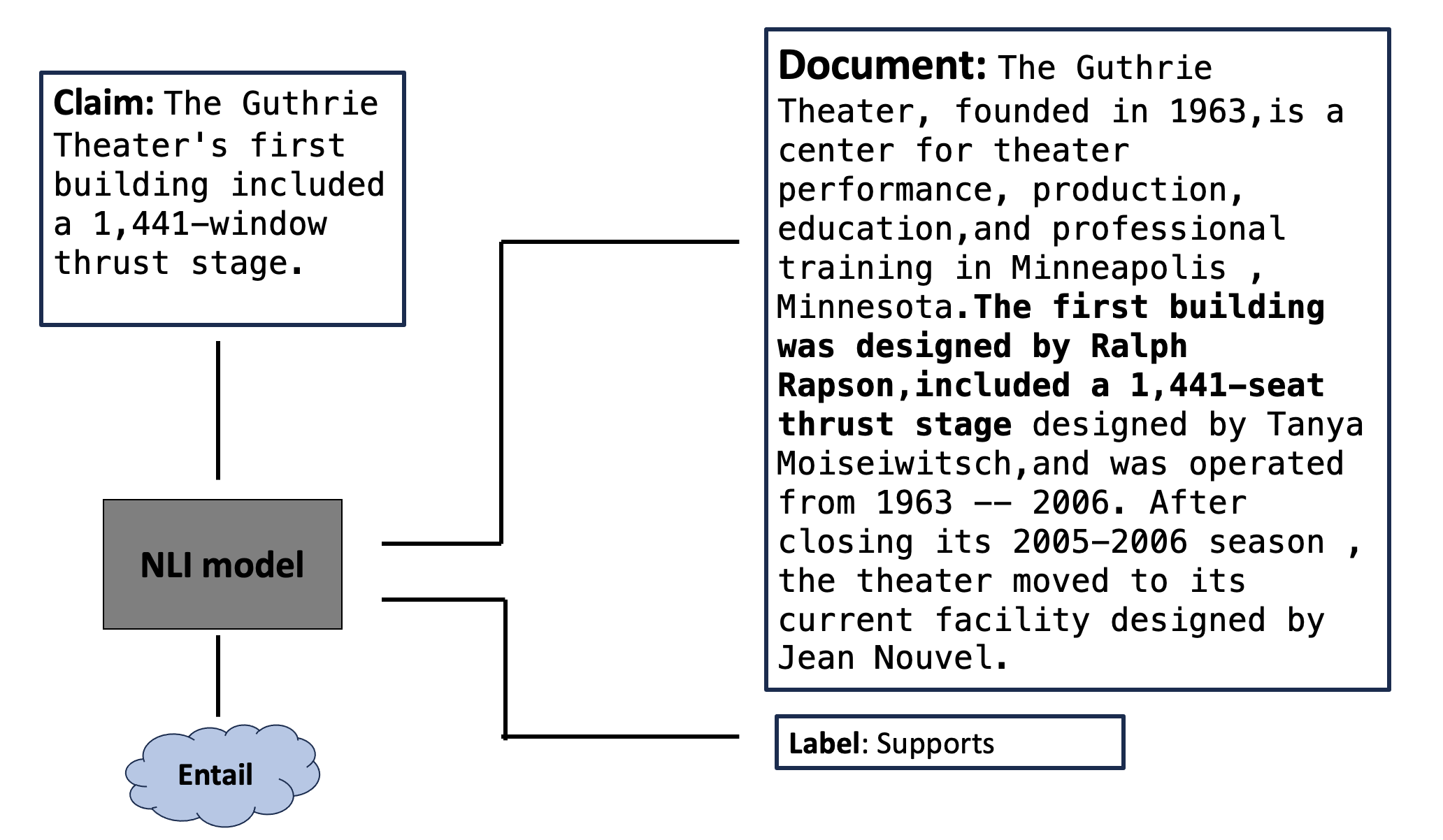} 
\caption{An example from the FEVER dataset, where the bold statement is the annotated rationale. Given the document and claim, the label denotes that the document contains evidence supporting the claim. The NLI predictor interprets this as a form of entailment between the claim and rationale.}
\label{fig:FEVER example}
\end{figure}

This differs from post-hoc techniques such as LIME~\cite{ribeiro2016should} or SHAP~\cite{lundberg2017unified}, specifically tailored to interpret black-box models. Although these techniques are model-agnostic by design, they are computationally expensive and do not guarantee faithfulness nor optimized for plausibility. Chain-of-thought (CoT)~\cite{wei2022chain} is another popular approach, aimed at prompting Large Language Models (LLM) such as OpenAI's GPT4 to elucidate its own prediction, in the form of reasoning steps which is said to be a form of explanation. However, we note that though the reasoning steps are seemingly plausible and convincing, there is no guarantee of the reasoning being faithful towards the supported output, since there is no constraint on the conditional variables. ETP models instead constrain the predictions on a compressed subset of the input, referred to as rationales, thereby guaranteeing the output to be solely conditioned on the subset, analogous to a binary form of feature relevance. 

In our work, we focus on improving the plausibility of rationales, measured via matching human annotations. Several work has established benchmark datasets that consist of both the task label as well as human-annotated rationales ~\cite{bao2018deriving,deyoung2019eraser}. Current works in extractive rationalization mostly implement a pipeline procedure of training an explainer and a predictor~\cite{deyoung2019eraser}, trained either jointly or separately. The training approach for these models can be bifurcated into two primary methods: supervised or unsupervised rationale extraction. In our methodology, we strike a balance by leveraging a minimal subset of annotated rationales ($\leq10\%$) to refine an ETP model. This refinement is applied to a separate NLI predictor, functioning as an auxiliary instructor for the explainer in the event of limited annotated rationales. More importantly, the explainer has no access to the annotations, which are exclusively presented to the NLI predictor.

Our approach is inspired by recent work in ensuring factual consistency in abstraction summarization~\cite{roit2023factually}, which has been found useful in cases of hallucination.  Firstly, we create an augmented dataset based on a label transformation algorithm, based on the annotated rationales and NLI classes. This is used to provide further training on the NLI predictor to generate sentence-level rationale annotations.

NLI models are designed to determine whether a hypothesis contradicts, entails, or is neutral to a given premise. As such, they provide useful signals to align a given explanation to the answer produced by the predictor, as shown in subsequent experiments, this can have some desirable effects on the robustness of rationales~\cite{chen2022can}. An example is illustrated in Figure~\ref{fig:FEVER example} on a fact verification task, where the purpose of the rationale is to act as evidence to either support or refute the given claim. In summary, the three key contributions of this work are the following:
\begin{itemize}
    \item A simple yet effective approach that improves the plausibility and robustness of extracted rationales, while simultaneously improving task performance. The approach achieves competitive results against supervised models while outperforming unsupervised models by a large margin (>100\%).
    \item To the best of our knowledge, this is the first work to utilize an auxiliary NLI predictor to generate augmented labels for extractive rationalization.
    \item Our approach has low resource requirements, using models of <300M parameters, and a small set of human-annotated rationales.
\end{itemize}

\section{Methodology}

\subsection{Problem setting}
Given an input document consisting of $N$ sentences, $x_i = \{x_{i,1},x_{i,2},...,x_{i,N}\}$. The task objective can be decomposed into two steps, namely rationale extraction, and task prediction. An explainer, $g_\phi$ takes in the input document and generates a binary mask over the sentences indicating the rationales, $g_\phi(\hat{z_i}|x_i) \in \{0,1\}_N$. 

The predictor, $f_\theta$ can only consider the masked inputs during inference, since the initial reason for extractive rationalization is to present the rationales as a faithful explanation towards the task prediction, $\hat{y_i} = f_\theta(\hat{z_i} \odot x_i)$, $ \odot $ is the element-wise multiplication. As rationales are designed to be a concise representation of the original text, there naturally exists a trade-off between generating a sparse $z$ and retaining sufficient information to accurately infer the task label. In various studies, optimization strategies are generally consistent, differing mainly in the use of human-annotated labels for training rationale extractors. Our approach, instead employs a semi-supervised method using an auxiliary predictor optimized for NLI, denoted as $f_{NLI}$.

\subsection{Semi-supervised NLI signal}
\begin{figure*}[t]
\centering
\includegraphics[width=0.7\textwidth]{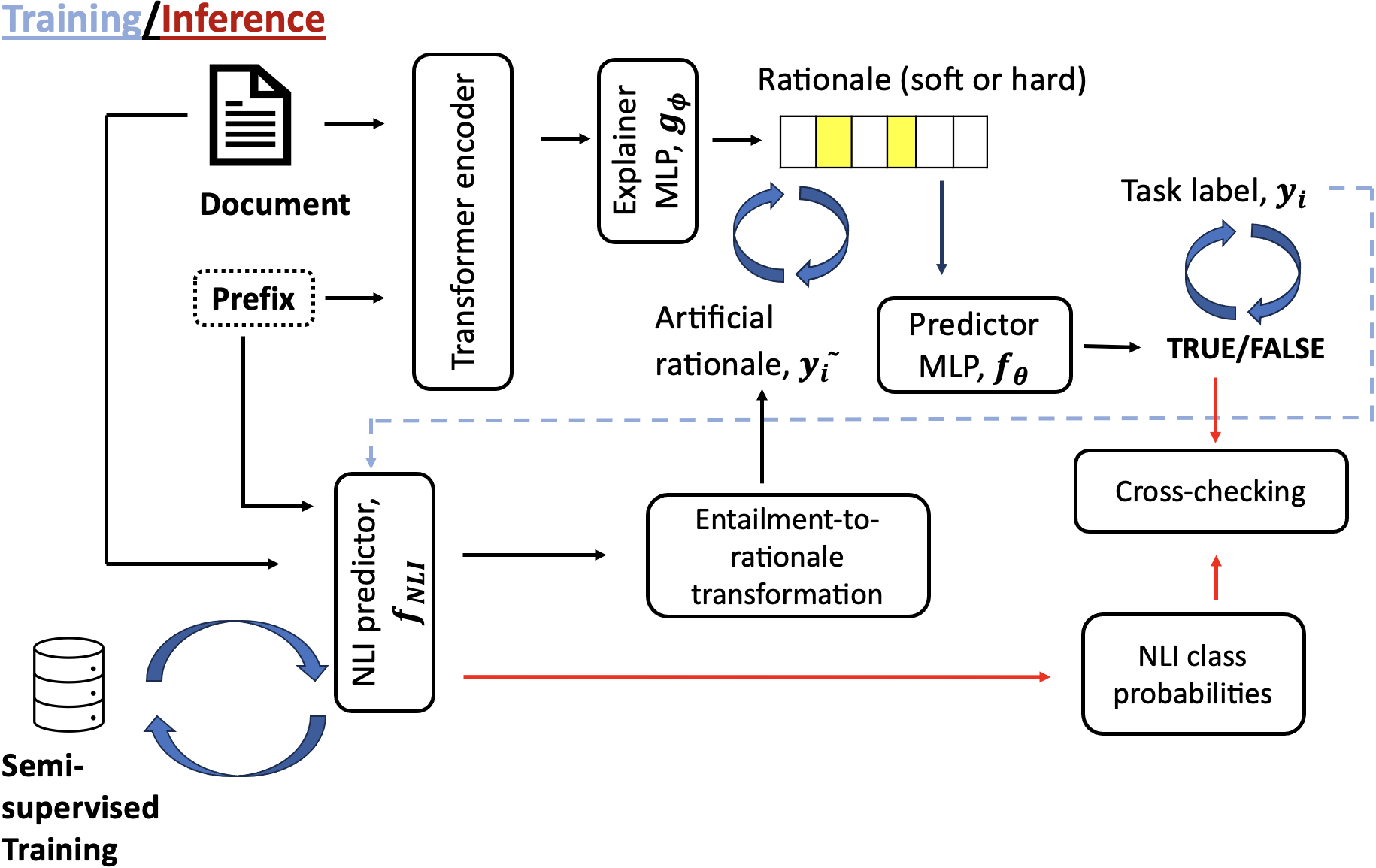} 
\caption{An overview of the proposed approach during training (bold in blue) and inference (bold in red). The NLI predictor only has access to the task label during training. The NLI predictor is initially fine-tuned using a limited set of annotated rationales, before generating artificial targets for the explainer. Cross-checking alignment is conducted during inference against the predictor (explained in Section~\ref{sec:inference}) }
\label{fig:overview}

\end{figure*}
Humans tend to prefer explanations that are aligned with the supported answer, similar to how NLI tasks involve generating the alignment between two sentences. As such, NLI predictors naturally serve as helpful supervision in the absence of annotated rationales. This is especially applicable in a fact-verification scenario where the task is to infer if a given claim is supported by the provided document. For example, given a document containing the following annotated rationale: \textit{"Kung Fu Panda opened in 4,114 theaters, grossing \$20.3 million on its opening day"} along with a claim: \textit{"Kung Fu Panda made more than \$1 million on opening day."}. The rationale acts as supporting evidence if the corresponding label, $y_i$ = \textit{SUPPORT}, indicates that the claim should be supported given the document and vice versa. The NLI predictor is fine-tuned based on this simple heuristic, to match each sentence in the document against the query. It is trained on the augmented dataset created via a label transformation technique shown in  Algorithm~\ref{alg:NLI}\footnote{Shown on FEVER dataset, where the supporting facts either support or refute the claim. Similarly applied for true/false settings.}. The transformation operates under the assumption that there are no contradictory rationales against the label, ie in $z_{i,j}$ contradicts the claim when the label is \textit{support}. This is a valid assumption since a sample with rationales contradicting the label should be considered an erroneous sample. The human- annotated rationales are used solely in the transformation, thus the explainer does not explicitly see the annotations.

During training, the NLI predictor acts as the source of supervision in place of the human-annotated rationales. As the explainer is trained to predict a binary mask, Algorithm~\ref{alg:NLI} can be implemented in reverse to transform the NLI outputs back to rationale labels, $\tilde{z}$ for the explainer's training, (see Appendix for more details). We note that the above approach is likewise applicable to any binary true/false tasks where the predictor has to indicate if the answer is true or false concerning the question.
\begin{algorithm}[tb]
\caption{Rationale to NLI label transformation}
\label{alg:NLI}
\textbf{Input}: Annotated rationale, $z_i$, task label, $y_i$\\
\textbf{Output}: NLI label,$\tilde{z_i}$
\begin{algorithmic}[1] 
\FOR{$z_{i,j}$ in $z_i$}
\IF {$z_{i,j} = 1 \land y_i$ = SUP}
\STATE $\tilde{z_{i,j}}$ = entailment
\ELSIF{$z_{i,j} = 1 \land y_i$ = REF}
\STATE $\tilde{z_{i,j}}$ = contradiction
\ELSE
\STATE $\tilde{z_{i,j}}$ = neutral
\ENDIF
\ENDFOR
\STATE \textbf{return} $\tilde{z_i}$ 
\end{algorithmic}
\end{algorithm}

\subsection{Sentence-level training}
We utilizes a pipeline approach consisting of a shared encoder, along with separate decoders for the explainer and predictor. The input is first encoded into contextualized hidden states, $h_{i,1:L} = enc(x_{i,1:L})$, where L is at the token level. We follow ~\cite{paranjape2020information} and transform the token-level hidden states into sentence-level by concatenating hidden states at the starting and ending positions and feeding it into an explainer to produce rationales, $\tilde{z_i} = g_\phi(h_i)$, where $h_i$ = $MLP(h_{i,s} \oplus h_{i,e})$, $ \oplus $ is the concatenation process. 

The predictor is conditioned on the rationales and trained using standard cross entropy.
\begin{equation}
\label{eq:ce}
    L_{f_\theta} = - \mathbb{E}_{z \sim g_\phi(z|x)} [\log (\hat{y}_i|\hat{z} \odot x)]
\end{equation}
The explainer loss, $L_{g_\phi}$ is similarly computed with (\ref{eq:ce}), but against the augmented targets, $\tilde{z_i} = f_{NLI}(\tilde{z_i}|x_i,y_i) \in \{0,1\}^N$, instead of the annotated targets. The full training and inference approach is depicted in Figure~\ref{fig:overview}, where the NLI predictor is first fine-tuned before training the ETP model. The choice of a shared encoder allows for a form of dependency between $e_i$ and $\hat{y_i}$, as the encoder has to jointly optimize the representation to infer both the task label and rationales accurately. The final loss is thus a combination of both the predictor and explainer cross-entropy loss, $L_{total} = L_{f_\theta} + \lambda L_{g_\phi}$, where $\lambda$ balances the trade-off between classification and plausibility performance.

The label transformation is only used during training as it requires access to $y_i$ which is not available at test time. However, we will show how $f_{NLI}$ can remain useful during inference.

\subsection{Inference}\label{sec:inference}
During inference, the rationales are extracted solely by the trained explainer, $f_\theta$. 
However, $f_{NLI}$ can act as a cross-checker against the predictor $g_\phi$ in the event of a distributional shift in $g_\phi$. Given $\hat{z_i}$ and a prefix (claim in fact verification or question-answer pair in Q\&A task), $f_{NLI}$ denotes if $\hat{z_i}$ contradicts or entails the prefix. We ignore the neutral class and re-weight the NLI class probabilities, $p(\tilde{y}^C_i)$ before averaging across the $n$ selected sentences in each instance,
\begin{equation}
    p(\tilde{y}^C_i) = \frac{1}{n} \sum_{j=1}^{n} p(\tilde{y}^C_{i,j})
\end{equation}
where C denotes the NLI class instance. The task probabilities, $p(\hat{y}_i^C)$ are then scaled with the NLI probabilities. 
\begin{equation}
    p(\hat{y}^C_i) = p(\hat{y}_i^C) \cdot p(\tilde{y}^C_i)
\end{equation}
This is helpful in the case where $f_{\theta}$ is less confident around the decision boundary and $f_{NLI}$ can provide additional support, given the additional training which aligns $f_{NLI}$'s decision between supporting rationales and task label.

\section{Experiments}
\label{experiment}
\subsection{Datasets}
We evaluate our approach against unsupervised and supervised baselines across three benchmark tasks from ERASER. ERASER contains a suite of NLP tasks, extended with human-annotated rationales, to assess plausibility.
\begin{itemize}
    \item \textbf{FEVER}: A fact-verification dataset, each instance consists of a claim and a document, where the goal is to determine if the claim is supported or refuted using information from the document. 
    \item \textbf{BoolQ}: Question-answering task, containing a context document from Wikipedia and a question, the answer is either true or false. Due to the long sequence, we select the most relevant portion of the context using TF-IDF scoring similar to ~\cite{paranjape2020information}.
    \item \textbf{MultiRC}: A multi-hop dataset, requiring reasoning over multiple sentences to infer to correct answer. Multiple answer choices can be associated with a single question and the task is to predict if the answer is true or false.
\end{itemize}

\subsection{Experimental Setup}
We use RoBERTa-base~\cite{liu2019roberta} as the shared encoder between the explainer and predictor. The NLI predictor, $f_{NLI}$ is a DeBERTa-large transformer~\cite{he2021debertav3} fine-tuned on multiple NLI datasets, we use the v3 variant. Our approach is agnostic to the choice of the pre-trained transformer for both the backbone encoder and NLI predictor. We selected RoBERTa-base, with its 125M parameters, due to its computational efficiency compared to larger models, while still maintaining high performance. We fine-tune the NLI predictor with 10\%\footnote{The restriction in training samples applies only to rationales, where we train the predictor on the full set of task labels.} of the annotated rationales. We list the full hyperparameter details in \ref{sec:hyperparameters}. A notable benefit of our approach is that it does not require an expensive search over objective-related hyperparameters.

\subsection{Baselines}
We evaluate our approach against both supervised and unsupervised settings, along with predictors subjected to full context. We refer to \textbf{Full-C} as the predictor-only set up to assess the gap in task performance between using the full context as compared to a subset. \textbf{Supervised} trains the explainer against human-annotated labels, $z_i$, instead of $\tilde{z_i}$ in our approach, serving as the upper bound for plausibility.

\textbf{IB} is an unsupervised approach from ~\cite{paranjape2020information} which optimizes a information-bottleneck objective and selects top $N\%$ according to pre-defined sparse prior. The author additionally introduces a semi-supervised approach of using 25\% of the annotated rationales which we refer to as \textbf{IB-25\%}. Note that this baseline is subjected to higher supervision compared to ours (10\%). We included the reported results for the sake of fairness \textbf{(R)}. We choose $10\%$ based on empirical results, serving as a good trade-off between minimal resource requirement and performance, albeit a comparable level of supervision ($25\%$) can be referred from Table~\ref{tab:ablation: size}. All evaluated approach implements an ETP-type setup, consisting of an explainer and predictor except for Full-C. 

\subsection{Metrics}
We report task performance using classification metrics such as accuracy and F1-score, while the plausibility of extracted rationales is assessed using F1-score~\cite{deyoung2019eraser} at the sentence level, Sentence-F1. We leave out any faithfulness metrics such as sufficiency as we assume ETP models to be inherently faithful given that the predictor is only subjected to the extracted explanation. We also assess the robustness by exposing the explainer to adversarial inputs~\cite{chen2022can} in Section~\ref{sec:robustness}.

We employ the following equations~\cite{chen2022can} to compute the normalized discrepancy in task performance, $\Delta_{T}$ and plausibility, $\Delta_{P}$ between the original and perturbed inputs as an indicator of robustness. Additionally, we utilize the attack rate, $AR$ to gauge the frequency with which the explainer identifies adversarial sentences. 
\begin{equation}
\label{eq:4}
    \Delta_{T} = \frac{1}{N}\sum_{i=1}^N \frac{M_{t}(\hat{y}_i,y_i) - M_{t}(\hat{y}_i^A,y_i)}{M_{t}(\hat{y}_i,y_i)}
\end{equation}
\begin{equation}
\label{eq:5}
    \Delta_{P} = \frac{1}{N}\sum_{i=1}^N \frac{M_{p}(\hat{z}_i,z_i) - M_{p}(\hat{z}_i^A,z_i)}{M_{p}(\hat{z}_i,z_i)}
\end{equation}
\begin{equation}
\label{eq:6}
    AR = \frac{1}{N}\sum_{i=1}^N \hat{z_i} \cap z^{AS}
\end{equation}
$M_t$ and $M_p$ is the scoring function for task and plausibility performance, for which we use the F1 and Sentence-F1 measurement. $\hat{y}_i^A$, and $\hat{z}_i^A$ refer to the generated class label and rationale given the adversarial input. $z^{AS}$ refers to the position of the adversarial prefix.

\begin{table*}[ht]
    \centering
    \small
    \begin{tabular}{l|c|c|c|c|c|c|c|c|c|c|c|c|c|c}
    \toprule
    & \multicolumn{3}{c|}{FEVER} & \multicolumn{3}{c|}{MultiRC} & \multicolumn{3}{c}{BoolQ} \\
    & \multicolumn{2}{c|}{Task} & Plausibility & \multicolumn{2}{c|}{Task} & Plausibility & \multicolumn{2}{c|}{Task} & Plausibility \\
    Approach & Acc & F1 & Sent-F1 & Acc & F1 & Sent-F1 & Acc & F1 & Sent-F1 \\
        \midrule
        \multicolumn{1}{l}{Full-C} & \multicolumn{1}{c}{93} &\multicolumn{1}{c}{ 91.8} & \multicolumn{1}{c}{-} & \multicolumn{1}{c}{\textbf{76}} & \multicolumn{1}{c}{\textbf{72}} & \multicolumn{1}{c}{-} & \multicolumn{1}{c}{65.8} & \multicolumn{1}{c}{53} & \multicolumn{1}{c}{-} \\
        \multicolumn{1}{l}{Supervised} & \multicolumn{1}{c}{90.1} &\multicolumn{1}{c}{ 88.4} & \multicolumn{1}{c}{\textbf{83.4}} & \multicolumn{1}{c}{74.3} & \multicolumn{1}{c}{70.5} & \multicolumn{1}{c}{\textbf{64.1}} & \multicolumn{1}{c}{\textbf{72.4}} & \multicolumn{1}{c}{\textbf{65.9}} & \multicolumn{1}{c}{\textbf{76}} \\
        \multicolumn{1}{l}{IB} & \multicolumn{1}{c}{85.9} &\multicolumn{1}{c}{85.9} & \multicolumn{1}{c}{38.9} & \multicolumn{1}{c}{64.1} & \multicolumn{1}{c}{63} & \multicolumn{1}{c}{23.1} & \multicolumn{1}{c}{64} & \multicolumn{1}{c}{63.5} & \multicolumn{1}{c}{10.3} \\
        \multicolumn{1}{l}{IB w 25\% } & \multicolumn{1}{c}{85.1} &\multicolumn{1}{c}{85.1} & \multicolumn{1}{c}{38.4} & \multicolumn{1}{c}{67.6} & \multicolumn{1}{c}{67.5} & \multicolumn{1}{c}{52.7} & \multicolumn{1}{c}{58.6} & \multicolumn{1}{c}{52.1} & \multicolumn{1}{c}{11.4} \\
        \multicolumn{1}{l}{IB w 25\% (R)} & \multicolumn{1}{c}{-} &\multicolumn{1}{c}{88.8} & \multicolumn{1}{c}{63.9} & \multicolumn{1}{c}{-} & \multicolumn{1}{c}{66.4} & \multicolumn{1}{c}{54} & \multicolumn{1}{c}{-} & \multicolumn{1}{c}{63.4} & \multicolumn{1}{c}{19.2} \\
        \midrule
        \multicolumn{1}{l}{Ours (10\%)} & \multicolumn{1}{c}{\textbf{93.7\textsubscript{+0.7}}} &\multicolumn{1}{c}{\textbf{92.6\textsubscript{+0.7}}} & \multicolumn{1}{c}{\textbf{80.1}} & \multicolumn{1}{c}{72.5\textsubscript{+0.2}} & \multicolumn{1}{c}{68.6\textsubscript{+0.5}} & \multicolumn{1}{c}{\textbf{56.4}} & \multicolumn{1}{c}{67.4\textsubscript{+2.2}} & \multicolumn{1}{c}{51.4\textsubscript{+9}} & \multicolumn{1}{c}{\textbf{29.6}} \\
        \bottomrule
    \end{tabular}
    \caption{Classification and plausibility performance comparison across the three ERASER tasks. Test results are averaged across 3 seeds. The subscript refers to the case where the NLI predictor is used as a cross checker, in \ref{sec:inference}. Results highlighted in bold refer to the best-performing approach. The supervised approach acts as the upper bound on plausibility performance. \textbf{R} is the reported results of the IB approach~\cite{paranjape2020information}.}
    \label{tab:1}
\end{table*}

\begin{table}[ht]
\centering
\begin{tabular}{ccc}
\toprule
FEVER & MultiRC & BoolQ \\
\midrule
100 & 56.7 & 20 \\
\bottomrule
\end{tabular}
\caption{Percentage of extracted over target rationales. BoolQ has the lowest percentage out of all three datasets.}
\label{tab:percentage}
\end{table}

\section{Results}
All results are averaged over three runs with different seeds. For Full-C, we do not report plausibility performance since there is no explainer module. In the ERASER benchmark, the number of annotated rationale varies between tasks, where BoolQ features a higher quantity of annotated sentences and also includes larger number of contiguous spans. The main objective of this study is to assess between different unsupervised and supervised approaches in generating plausible and robust rationales, while minimizing negative effects on downstream task performance.

\subsection{Plausibility and Task Analysis}
The task and plausibility performance is shown in Table~\ref{tab:1}. Judging from the results, our approach achieves highly competitive performance against the gold standard for both task (Full-C) and plausibility (Supervised). In FEVER, it even surpasses the full context approach (94.2 vs 93). It goes to show that ETP-like models can benefit from ignoring spurious noise by conditioning the predictor to only text considered essential for inferring the target class. The additional usage of $f_{NLI}$ as a cross-checker during inference also provided considerable improvements across all three benchmarks, at little to no cost in computational resources.

In terms of plausibility, our method delivers a plausibility score that is on par with the fully supervised approach across all datasets except BoolQ. We note that a likely reason is that the target rationales are largely inconsistent in length, with instances stretching across as many as six contiguous sentences. Since the NLI predictor is optimized toward matching each sentence with the given query. It may fare worse when individual sentences appear to be unrelated to the query but are nonetheless annotated as rationales. Table~\ref{tab:percentage} shows the percentage proportion of sentences annotated as rationales over the target. It's noteworthy that the explainer marks fewer sentences due to the NLI predictor's tendency to classify the majority of sentences as neutral, deeming them non-essential for task prediction.

Nonetheless, optimizing the explainer with NLI supervision is proven to be superior compared to the unsupervised information bottleneck objective by ~\cite{paranjape2020information}. Our approach outperforms the former by large margins in terms of plausibility on FEVER ($>25\%$) and BoolQ ($>50\%$), even when provided with a lower amount of supervision (10\% vs 25\%). The performance gap is even larger when compared to fully unsupervised (IB), with more than twice the scores. The IB method learns a sparse mask over the input document, $x_i$ by maximizing the mutual information between rationale $z_i$ and task label $y_i$ while limiting the extraction budget to a pre-defined prior. However, estimating the prior is difficult and can be detrimental in instances with varying rationale lengths such as in BoolQ. Our approach sidesteps the complicated training yet achieves a better-tuned explainer in extracting plausible rationales.

\subsection{Robustness}
\label{sec:robustness}
In this section, we evaluate the robustness of ETP models when faced with inputs prefixed with an adversarial query. The query is unrelated to the document and carries a contrastive meaning with respect to the original. For example, given a claim in FEVER, \textit{"Earl Scruggs was a musician who played banjo."}, the noun, \textit{"Earl Scruggs"} and \textit{"banjo"} is replaced to form the adversarial sentence \textit{"manchester archer was a songwriter who played mandolin ."}. The attack is minimally changed from the query to distract the explainer. A model with limited robustness might interpret the attack as pertinent due to its analogous semantics, thereby influencing the predictor and undermining task performance. The robustness results are reported in Table~\ref{tab:robustness}. We found similar findings as compared to ~\cite{chen2022can} who note that ETP models exhibit greater robustness compared to predictors subjected to the full context. 

In the FEVER dataset, our approach suffers the lowest drop in task and plausibility performance, while having the lowest AR in both datasets. IB has the highest AR, even extracting every adversarial sentence in FEVER. A contributing factor to our approach's low AR rate is that the NLI signal is derived by verifying if a sentence aligns with the query based on the provided task label. This strengthens the explainer's proficiency in dismissing instances that don't satisfy this criterion. On the other hand, the explainer trained with IB is emphasized to maximize the task objective, which can lead to situations where a minimally perturbed sentence is mistakenly perceived as useful. This further proves that training with NLI feedback produces more robust and plausible models.

\begin{table}[ht]
\centering
\small
\begin{tabular}{l|ccc|ccc}
\toprule
& \multicolumn{3}{c|}{FEVER} & \multicolumn{3}{c}{MultiRC} \\
Approach & $\Delta_{T}$ & $\Delta_{P}$ & AR & $\Delta_{T}$ & $\Delta_{P}$ & AR\\
\midrule
Full-C &  11.2&  - & - & 29.6  & - & -\\
Supervised & 10.8 & 37 & 54.6 & 14.3 & 26.7 & 68.3\\
IB (25\%) & 12 & 35.8 & 100 & \textbf{4.9} & \textbf{19.1} & 93\\
Ours (10\%) & \textbf{7.8} & \textbf{8.4} & \textbf{32.3} & 10.2 & 27.1 & \textbf{67.2}\\
\bottomrule
\end{tabular}
\caption{In both FEVER and MultiRC, we measure robustness with a preference for lower values. Models considering the full context are evaluated solely based on the difference in task performance as they don't engage in rationale extraction. All values are normalized percentages drop computed via (~\ref{eq:4}),(~\ref{eq:5}) and (~\ref{eq:6})} 
\label{tab:robustness}
\end{table}

\section{Ablation}
\subsection{Importance of NLI training}
In this study, we seek to question the usefulness of introducing further fine-tuning using the limited set of annotations. While the NLI predictor is previously fine-tuned on various NLI tasks, the sentence lengths in its training distribution differ from those in our experimental datasets. Additionally, domain-specific semantics differences can introduce variations in the NLI predictor's inference process. 
Consequently, the NLI predictor might not always accurately discern the NLI class, leading to the generation of misleading signals for the explainer. To quantify the effectiveness of further fine-tuning, we compute the drop-in performance on both task and plausibility between an NLI predictor that is fine-tuned, referred to as \textbf{FI} and one that is not, \textbf{NFI}.
\begin{figure}[t]
\centering
\includegraphics[width=\columnwidth]{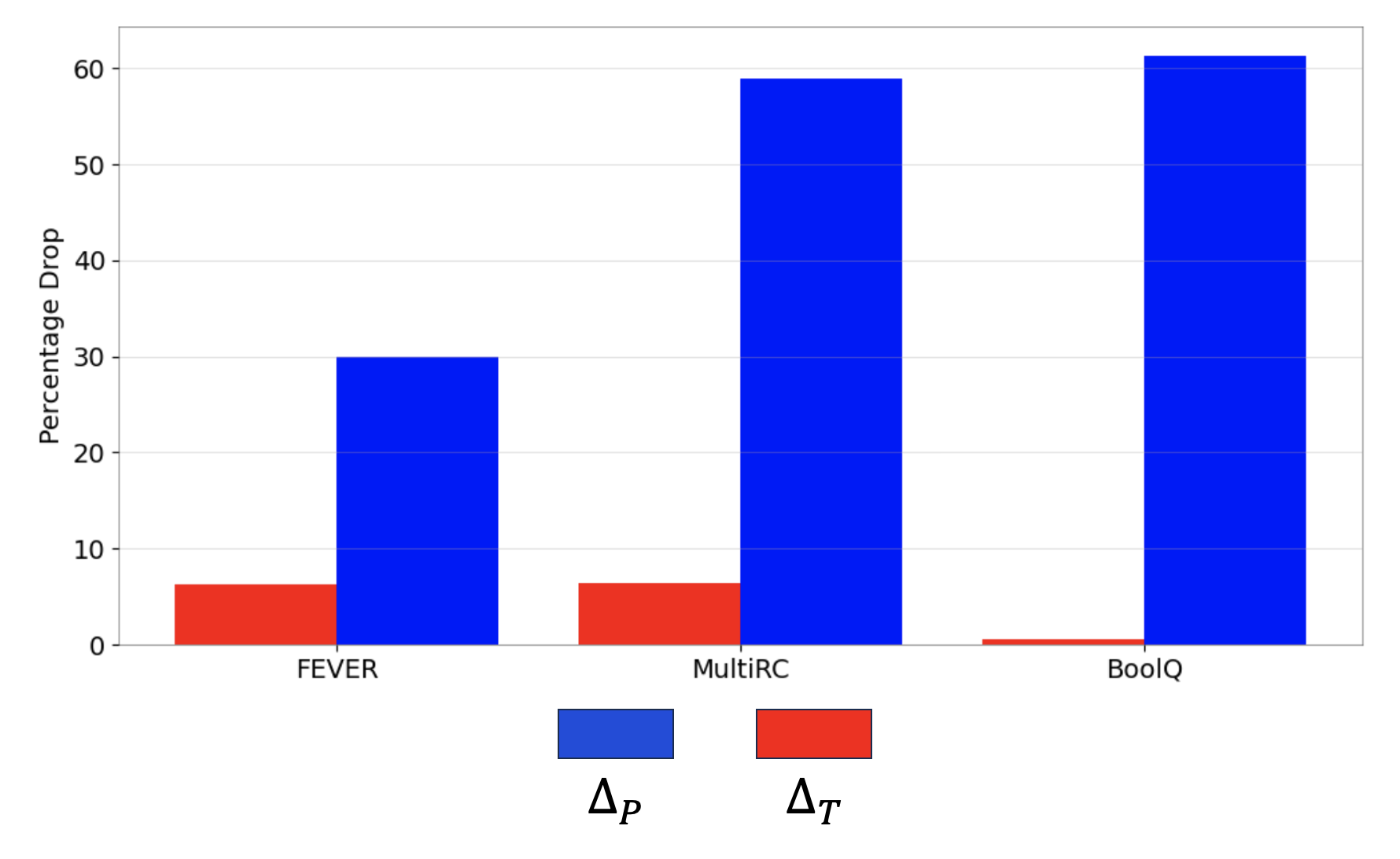} 
\caption{Task and Plausibility performance drop when there is no further fine-tuning on the NLI predictor (10\% data). The metrics are computed similarly to robustness using (~\ref{eq:4}) and (~\ref{eq:5}) and are presented in normalized percentages.}
\label{fig:ablation}
\end{figure}
\iffalse
\begin{table}[ht]
\centering
\begin{tabular}{cccc}
\toprule
Metrics & FEVER & MultiRC & BoolQ \\
\midrule
$\Delta_{T}$ & 6.2 & 6.4 & 0.6 \\
$\Delta_{P}$ & 30 & 58.9 & 61.3 \\
\bottomrule
\end{tabular}
\caption{Task and Plausibility performance drop when there is no further fine-tuning on the NLI predictor (10\% data). The metrics are computed similarly to robustness using (~\ref{eq:4}) and (~\ref{eq:5}) and are represented in normalized percentages.}
\label{tab:ablation}
\end{table}
\fi
The gap in task and plausibility performance is reported in Figure~\ref{fig:ablation}.

\begin{figure}[t]
\centering
\includegraphics[width=\columnwidth]{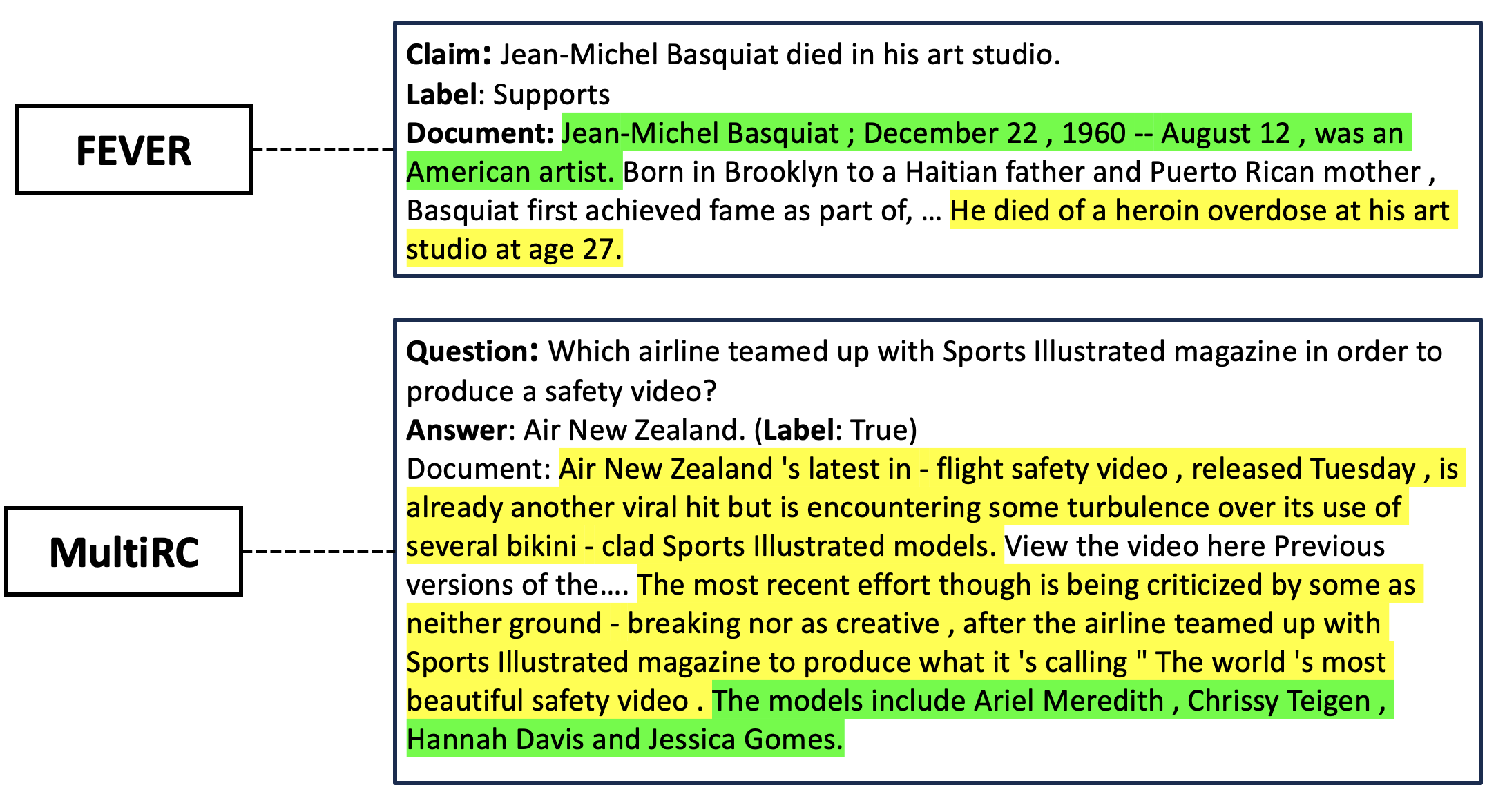} 
\caption{Example of query and input document where the sentences highlighted in green refer to the NLI predicator without fine-tuning. Yellow refers to the annotated rationale as well as extracted by the fine-tuned predictor.}
\label{fig:example}
\end{figure}

These results substantiate our initial hypothesis. Without fine-tuning, NFI struggles to provide meaningful feedback to the explainer, primarily because of its limited capability to accurately determine whether a specific sentence should support or contradict the query based on the given task label. Taking a closer look at sentence classifications in Figure~\ref{fig:example} reveals that the NFI tends to mistakenly identify neutral sentences as entailments. In the FEVER example, although the initial sentence shares a noun with the claim, it does not address the death of the noun's subject yet the NFI incorrectly recognizes it as entailment. 
Similarly, in the MultiRC instance, the concluding sentence lacks any significant connection to the given question or answer. This may be the reason why despite a significant drop in accurately extracting the correct rationale, $58.9\%$ in MultiRC and $61.3\%$ in BoolQ, the task performance surprisingly does not incur a huge loss ($<10\%$). A neutral sentence would not drastically change the class probabilities of the predictor as compared to a contradicting sentence. Nevertheless, incorporating additional fine-tuning on the NLI predictor is still essential in filtering out false positives such as sentences with neutral relationships in inferring the output.

\subsection{Model sizes and NLI supervision}
\begin{table}[ht]
\centering
\small
\begin{tabular}{l|ccc}
\toprule
Ablation type & Acc & F1 & Sent-F1\\
\midrule
Original (base w 10\%) &  72.5&  68.6 & 56.4\\
Large & 74.3 & 71.2 & 58.1 \\
Base w 25\% & 73.3 & 71.1 & 57.9 \\
Base w 50\% & 73.8 & 70.9 & 58 \\
Without NLI Pre-FT & 69.2 & 64.3& 52.6\\
\bottomrule
\end{tabular}
\caption{Ablation on model size, \% NLI supervision and effects of not doing pre-finetuning of $f_{NLI}$ on general NLI tasks (SNLI, MultiNLI). Implemented on MultiRC.} 
\label{tab:ablation: size}
\end{table}
We carry out further analysis on the effect of both model size and amount of NLI supervision given to $f_{NLI}$. We compare RoBERTA-large (330M) with the original $10\%$ supervision and the base model with increased level of supervision $\in [25,50]$. We additionally compare a DeBERTa encoder without prior fine-tuning on NLI datasets, while similarly fine-tuning on 10\% of annotated rationales. The benefits of using a larger encoder and increased NLI supervision for $f_{NLI}$ can be observed from Table~\ref{tab:ablation: size}. Notably, there is little difference in both accuracy and plausibility scores with higher supervision. Furthermore, our approach remains effective using off-the-shelf encoders without prior fine-tuning. This highlights the strength of our approach which remains effective even in low-resource conditions. 

\section{Related Works}
\textbf{Linguistic Interpretability}: In recent years, extractive rationalization and attention-based interpretability have emerged as significant approaches within AI research, with numerous studies contributing to the field~\cite{gurrapu2023rationalization, mohankumar2020towards, serrano2019attention}. However, the efficacy of using attention as an interpretability mechanism has been debated, highlighting a division among researchers. Critics argue that attention scores do not significantly impact model predictions and present challenges in generating counterfactuals~\cite{jain2019attention}, and may be biased due to their reliance on neighboring token information~\cite{bai2021attentions,tutek2022toward}. In contrast, proponents suggest that the relevance of attention weights varies with the definition of faithfulness and that multiple weight combinations can yield the same output~\cite{wiegreffe2019attention}. Efforts to enhance the reliability of attention mechanisms include task-specific attention constraints~\cite{chrysostomou2021improving} and penalties on scores for key words~\cite{chrysostomou2021enjoy}.\\\\
\textbf{Extractive Rationalization}: The foundational work by ~\cite{lei2016rationalizing} introduced extractive rationalization using REINFORCE with sparsity regularization for end-to-end training of explainers via predictors' objectives. The ERASER benchmark, established by ~\cite{deyoung2019eraser}, evaluates explainer-predictor models across seven NLP tasks, utilizing a BERT-to-BERT framework and sequential training. Subsequent research, such as ~\cite{atanasova2022diagnostics}, focused on explainer consistency and confidence, while ~\cite{lakhotia2020fid} employed Fusion-In-Decoder for rationale extraction in lengthy documents.

The challenge of acquiring supervised rationales has spurred interest in unsupervised methods for generating reliable rationales. Efforts include ~\cite{paranjape2020information}, targeting rationale conciseness through information bottleneck optimization, and ~\cite{ghoshal2022quaser}, which mitigates spurious correlations in QA by adding a question generation objective. ~\cite{jain2020learning} modularizes the objective, and ~\cite{glockner2020you} uses sentence-specific encoding with aggregated loss for rationale selection. Unlike these methods, our approach seeks to optimize all rational sentences without relying on the assumption of pre-defined rationale availability.\\\\
\textbf{NLI signals}: There have also been works utilizing NLI signals to enhance downstream tasks. ~\cite{roit2023factually} directly uses entailment scores as a reward signal to optimize factual summarization using RL,~\cite{laban2022summac,kryscinski2019evaluating} for mitigating inconsistency in abstractive summarization. ~\cite{chen2023zara} performs self-rationalization training using a small set of annotated rationales and then annotating the rest of the unlabelled dataset. However, the work is based on abstractive setting, using free-text self-generated rationales, thereby violating the faithfulness property of the explanation.~\cite{golovneva2022roscoe} utilizes NLI as a metric for ensuring semantic correctness in explanations.

\section{Conclusion}
In this paper, we have introduced a simple yet unique way of generating artificial learning signals from an alternative source, to cope with scenarios where human-annotated rationales are scarce. The method harnesses a transformer pre-trained on the NLI task. Through additional fine-tuning, the NLI predictor can produce less biased labels, enhancing the learning process for the explainer. 

Through the extensive experiments conducted, we have shown that our work can alleviate the plausibility and robustness of ETP models in a low-resource environment. Notably, with just $10\%$ of the annotated rationale, our method delivers performance on par with fully supervised models and significantly outperforms both semi-supervised approaches that utilize more annotated data and unsupervised settings. 
In future directions, we plan to extend this work toward models that generate abstractive explanations, where the NLI signal can act as verification feedback to ensure the mitigation of biased explanations. Another interesting direction is to study how can we extend the NLI predictor's coverage beyond a single sentence, to capture the correspondence between longer documents.

\section{Limitations}
We only evaluate a singular trait of interpretability: plausibility. We note that multiple other traits of interpretability are equally important and we leave that to further work. The sizes of the encoder models implemented in this work are relatively small, with the biggest consisting of ~300M parameters. Though model scaling is the primary objective, we note the importance of extending our work towards larger models given the popularity of NLP research surrounding LLMs.

\bibliography{custom}

\appendix
\section{Appendix}
In the main paper, we showed how the label transformation technique is used to transform an annotated rationale into an NLI-associated label, for the purpose of fine-tuning the NLI predictor. We will now show how the reverse is applied to facilitate the training of the explainer.

\subsection{Reverse label transformation}
Given a query, $q$ and each sentence, $x_i$, we concatenate the query and sentence as input to the NLI predictor, where the NLI class label is generated as $\tilde{y_i} = f_{NLI}(q \oplus x_i)$. This applies to both queries with a single sentence such as the claim in FEVER and BoolQ or double sentences in MultiRC, comprising of both the question and answer. The NLI class, $\tilde{y_i}$ is used together with the task label, $y_i$ to generate $\tilde{z_{i}}$, used in place of $z_i$ for the semi-supervised explainer. The transformation is detailed in Algorithm~\ref{alg:reverse}, applied in reverse to Algorithm~\ref{alg:NLI}. C refers to Contradiction, and E to Entailment (example shown on FEVER task).
Note that if the $f_{NLI}$ indicates that the sentence is neutral to the query, the sentence is automatically labeled as a non-rationale. This is similar in the case where if a document is annotated as false with respect to the query, all rationales should be a contradiction and vice versa.
\begin{algorithm}[H]
\caption{Reverse label transformation}
\label{alg:reverse}
\textbf{Input}: query, $q_i$, input document, $x_i$, task label, $y_i$ and NLI predictor, $f_{NLI}$\\
\textbf{Output}: NLI label,$\tilde{z_i}$ 
\begin{algorithmic}[1] 
\FOR{each $x_{i,j} \in x_i$}
\STATE $\tilde{y_i} \gets f_{NLI}(q_i \oplus x_{i,j})$
\IF {($\tilde{y_i} = \text{E}$ and $y_i = \text{SUP}$) or ($\tilde{y_i} = \text{C}$ and $y_i = \text{REF}$)}
\STATE $\tilde{z_{i,j}} \gets 1$
\ELSE
\STATE $\tilde{z_{i,j}} \gets 0$
\ENDIF
\ENDFOR
\STATE \textbf{return} $\tilde{z_i}$ 
\end{algorithmic}
\end{algorithm}

\subsection{Hyperparameters}
\label{sec:hyperparameters}
 We use the AdamW optimizer from ~\cite{loshchilov2017decoupled} with $\epsilon$ set at 1e-8 and fix the batch size at 8. We use a learning rate warm-up scheduler with the final rate capped at 2e-5 and clip all gradient norms at a value of 1.0 while applying a dropout of 0.2 for the explainer. The explainer is a two-layer MLP with ReLU activation. Early stopping is implemented where the training is stopped if the validation loss does not improve after 3 epochs. We run all our experiments for a maximum of 10 epochs, on NVIDIA A6000s, implemented with PyTorch. We do not find much difference between various values of $\lambda$ and set it to 1.
\end{document}